\documentclass[runningheads]{llncs}

\usepackage[mobile]{eccv}

\usepackage{eccvabbrv}

\usepackage{graphicx}
\usepackage{booktabs}
\usepackage{multirow}
\usepackage{subcaption}
\usepackage{booktabs}
\usepackage{colortbl}
\usepackage{arydshln}
\usepackage{wrapfig}

\usepackage[accsupp]{axessibility}  % Improves PDF readability for those with disabilities.

\usepackage[pagebackref,breaklinks,colorlinks,citecolor=eccvblue]{hyperref}
\usepackage{orcidlink}

\begin{document}

\title{RESTORE: Towards Feature Shift for Vision-Language Prompt Learning} 

\titlerunning{Abbreviated paper title}

\author{Yuncheng Yang\inst{1} \and
Chuyan Zhang\inst{1} \and
Zuopeng Yang\inst{2} \and Yuting Gao\inst{3} \and Yulei Qin\inst{3} \and Ke Li\inst{3} \and Xing Sun\inst{3} \and Jie Yang\inst{1} \and Yun Gu\inst{1}}

\authorrunning{F.~Author et al.}

\institute{Shanghai Jiao Tong University \\
\email{Yaphabates@sjtu.edu.cn}\\ \and
Guangzhou University\\ \and Tencent Youtu Lab}

\maketitle

\begin{abstract}
Prompt learning is effective for fine-tuning foundation models to improve their generalization across a variety of downstream tasks.
However, the prompts that are independently optimized along a single modality path, may sacrifice the vision-language alignment of pre-trained models in return for improved performance on specific tasks and classes, leading to poorer generalization.
Existing work does not take such drawbacks into serious consideration, for the simple reason that the evaluation of visual and textual alignment for general purposes is excluded in experiments.
In this paper, we first demonstrate that prompt tuning along only one single branch of CLIP (e.g., language or vision) is the reason why the misalignment occurs.
Without proper regularization across the learnable parameters in different modalities, prompt learning violates the original pre-training constraints inherent in the two-tower architecture.
To address such misalignment,
we first propose feature shift,
which is defined as the variation of embeddings after introducing the learned prompts,
to serve as an explanatory tool.
We dive into its relation with generalizability and thereafter propose RESTORE,
a multi-modal prompt learning method that exerts explicit constraints on cross-modal consistency.
To be more specific, in order to prevent feature misalignment,
a feature shift consistency is introduced to synchronize inter-modal feature shifts by measuring and regularizing the magnitude of discrepancy during prompt tuning.
In addition, we propose a "surgery" block to avoid short-cut hacking,
where cross-modal misalignment can still be severe if the feature shift of each modality varies drastically at the same rate.
It is implemented as feed-forward adapters upon both modalities to alleviate the misalignment problem.
Extensive experiments on 15 datasets demonstrate that our method outperforms the state-of-the-art prompt tuning methods without compromising feature alignment.
Codes and models are available at \url{https://github.com/Yaphabates/RESTORE_}.
\end{abstract}

\section{Introduction}
\label{sec:intro}

\begin{figure*}[!t]
	\centering
	\includegraphics[width=1.0\linewidth]{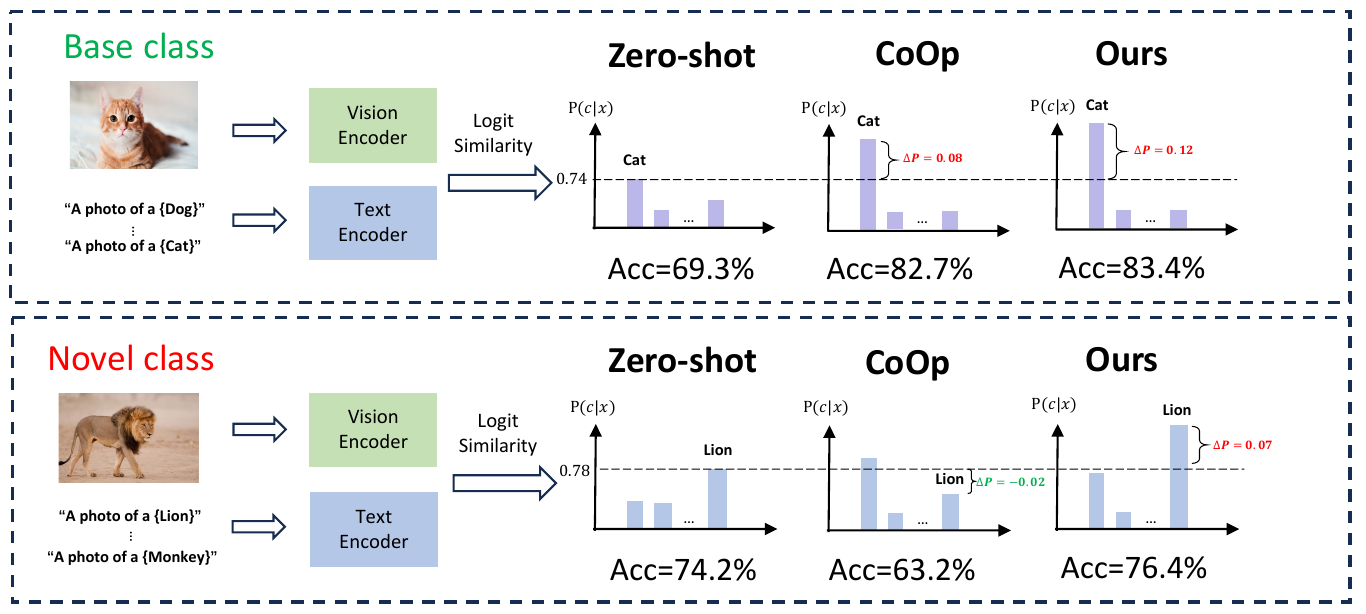}
	\caption{
            The pre-trained VLM (e.g., CLIP) demonstrates a strong generalizability
            % which is demonstrated by
            by its zero-shot performance on both base and novel classes.
            % the performance of zero-shot and the zero-shot model in the Fig.~can achieve good results on both CLIP on both base and new classes.
            However, existing prompt learning methods over-emphasize performance gains on the seen base classes while ignoring their declining generalization on novel ones, which is demonstrated by the decreased probability $P(c|x)$ of the ground-truth category and the overall accuracy.
	}
	\label{fig: illustraion_0}
\end{figure*}

Vision-Language Models (VLMs)~\cite{radford2021learning,jia2021scaling} have demonstrated remarkable generalization capabilities across multiple tasks.
The CLIP~\cite{radford2021learning} is pre-trained on 400 million pairs of images and texts with numerous computing resources,
and it is impractical to conduct full parameter fine-tuning of the entire model for downstream tasks due to the high cost of collecting manually annotated large datasets with similar scale.
% collecting \textit{large-scale datasets} with labels.
For efficient tuning with smaller datasets and fewer trainable parameters, \textbf{prompt tuning}~\cite{li2021prefix,jia2022visual,zhou2022learning}, \textbf{lightweight adapters}~\cite{li2023graphadapter,zhang2021tip,zhang2023prompt}, and \textbf{low-rank parametric bypass matrices}~\cite{hu2021lora} have been investigated recently.
As one of the most prospective fine-tuning techniques,
prompt tuning is proposed to efficiently adapt foundation models to downstream tasks by introducing learnable prompts while freezing the backbones~\cite{zhou2022conditional, zhou2022learning,liu2023gpt,lester2021power,zang2022unified,chen2022prompt,yao2023visual,guo2023texts,xing2022class}.
The learned prompts interweave pre-trained features via the attention mechanism in the vision or language branch,
enforcing task-specific distribution shifting.

Despite the improved performance on categories and tasks of interest,
recent studies~\cite{zhou2022conditional,khattak2023self} reveal that prompt tuning spoils the parameterized knowledge acquired in pre-training.
Without appropriate constraints,
the inherent cross-modal alignment is sacrificed for overfitting the downstream tasks, causing the loss of model generalization.
% It enforces distribution shifting towards the overfitting of downstream tasks
% the generalizability of pre-trained models degrades rapidly during prompt tuning.
% process, the generalization performance of the model decreases.
% For instance,
As shown in Fig.~\ref{fig: illustraion_0},
compared with the zero-shot CLIP,
the fine-tuned CLIP by CoOp~\cite{zhou2022learning} experiences a performance reduction by 11\% on unseen classes across 11 distinct datasets.
%In the case of CLIP,
%the extensive pre-training on large datasets equips it with a noise-robust embedding space where image and text representations are aligned for excellent generalization ability.
% due to the introduction of learnable prompts by CoOp,
% the performance of the CoOp on base classes is improved while that on novel classes deteriorates.
% implying that the generalizability is sacrificed in return for biased, task-specific overfitting.
% which means the generalization ability is devastated.
Therefore,
it is of great importance to simultaneously enhance downstream performance and maintain strong generalizability after prompt tunning.
Existing studies proposed implicit or explicit regularization to better preserve the general knowledge via various techniques like early stopping,
data augmentation~\cite{qin2021lfpt5,gao2020making,zhu2023prompt,li2023gradient},
gradient and prompt constraints~\cite{zhou2022conditional,khattak2023self}.
However, these methods fail to systematically analyze the fundamental reasons behind such model degeneration after prompt tuning.
%Specifically,
%\cite{zhou2022conditional} does not notice that such degradation is caused by feature misalignment of different modes,
%}
% However, \cite{zhou2022conditional} does not analyze the degradation of model generalization caused by misalignment of different modes,
%while \cite{khattak2023self} only focuses on the output embeddings of vision and language encoders without \qyl{delving deeper into the layer-wise statistics of backbone features.}
% modifying the embedding in the backbone.

% To better understand the underlying reason for the degradation of generalization,
In this paper, we provide a pioneering explanation of the underlying reasons behind the degradation of model generalization.
% in prompt tuning.
% we concentrate on \textbf{feature shift},
We propose the \textbf{feature shift} as a tool to quantify the inter-modal discrepancy of representations,
and thereafter discover the relationship between such discrepancy and model generalization.
% where the discrepancy of embeddings occurs inevitably due to the introduction of learnable prompts.
% Since the embedding brought by learnable prompts exists in different layers of the backbone network,
% it is necessary to align the feature differences brought about by the introduction of prompts in the encoder.
% Despite the fixed parameters of visual and textual encoders,
The newly incorporated learnable prompts,
either present in vision or language branches,
contribute to the cumulative feature shifts layer-by-layer.
Consequently,
asynchronous changes occur independently over features of the image and text modalities under the data-scarce scenarios of downstream fine-tuning.
% in the context of
% the encoders from the two modalities experience in their embeddings.
% Therefore, in this paper,
% through prompt tuning
We primarily address such inter-modal inconsistency and propose \textbf{RESTORE}: towa\textbf{R}ds f\textbf{E}ature \textbf{S}hif\textbf{T} for visi\textbf{O}n-language p\textbf{R}ompt l\textbf{E}arning.
Specifically,
we design a cross-modal constraint on feature shift,
which minimizes the measured distance between the visual and textual feature shifts to impose constraints on the synchrony of image and text encoders for alignment.
% discrepancy of feature shifts 
% feature shift values of both visual and linguistic modalities,
% requiring the feature shift values of the two modalities to be as close as possible.
In order to prevent the prompt-intervened features severely deviating from the pre-trained features simultaneously for both vision and language modalities,
we further develop a "surgery" block to act upon the output features for mitigation of misalignment.
% thereby mitigating the representation misalignment.
% Furthermore, we have integrated an additional 'surgery' block, 
Such blocks,
implemented as adapters,
are respectively governed by the extent of visual and textual feature shift and thereby correct representations dynamically.
Our main contribution can be summarized as follows:

\begin{enumerate}
\item We systematically and quantitatively explain the reason,
namely feature shift,
behind the degraded generalizability of VLMs during prompt tuning.
\item We propose the feature shift consistency loss from the perspective of cross-modal alignment 
% between different modalities of the VLM
to minimize 
% the distance embedding
the discrepancy between the two modalities.
\item We propose the "surgery" block to counteract potential severe feature shifts to avoid overfitting and ensure alignment on output representations.
% controlled by the severity of the feature shift to alleviate the feature shift and prevent models from collapsing.
% This "surgery" block prevents the model from misalignment problems and improves the performance on both seen and unseen tasks.
\item Extensive experiments of few-shot prompt tuning on 11 datasets
confirm the superiority and validity
% demonstrate the effectiveness 
of our feature shift loss and "surgery" block
% with superior performance
over state-of-the-art (SOTA) methods.
%The proposed method also brings supplementary benefits to existing methods like MaPLe\cite{khattak2023maple} and PromptSRC~\cite{khattak2023self}.
\end{enumerate}

\section{Related Work}
\subsection{Vision-Language Model}

Recently,
VLMs \cite{radford2021learning}\cite{jia2021scaling}\cite{zhai2022lit}\cite{yao2021filip}\cite{yuan2021florence}\cite{gao2022pyramidclip}\cite{gao2023softclip} have showcased remarkable achievements across a broad range of tasks.
The high transferability of pre-trained VLMs has been validated on
% to new tasks
downstream tasks such as few-shot and zero-shot image recognition ~\cite{gao2023clip}~\cite{zhou2022learning}, cross-modal generation~\cite{nichol2021glide}~\cite{patashnik2021styleclip}~\cite{ramesh2022hierarchical} and vision question answering~\cite{zhang2023pmc}~\cite{hu2023promptcap}.
% is impressive.
The VLM pre-training is usually guided by certain vision-language objectives~\cite{yao2021filip}~\cite{yu2022coca} that enable models to learn image-text correspondences from the large-scale image-text pairs~\cite{schuhmann2022laion}~\cite{zhang2023vision}.
Nonetheless, effectively applying VLMs to downstream tasks remains a complex issue. A typical model is CLIP~\cite{radford2021learning}, which utilizes vision-language contrastive learning for informative vision-language representation.
In the present study, we choose CLIP as our testbed due to its wide usability and comparability.

\subsection{Multi-Modal Prompt Tuning for VLMs}
Prompt tuning is first proposed in NLP to effectively fine-tune the large-scale model with the backbone frozen~\cite{li2021prefix}~\cite{lester2021power}.
The instructions in the form of a fixed sentence template are usually given to the language model (known as language prompt), allowing it to better understand the task. CoOp~\cite{zhou2022learning} proposed to replace the completely fixed template with a learnable prompt.
Co-CoOp~\cite{zhou2022conditional} added conditioning prompts on image instances to avoid overfitting to certain tasks.
Since both vision and language inputs are unified into the same structure under the transformer framework, similar to language prompt tuning, vision prompt also introduces a small number of learnable parameters into the input space while freezing the entire pre-trained transformer backbone during downstream training~\cite{jia2022visual}~\cite{huang2023diversity}~\cite{xing2022class} \cite{chen2023semantic}.
Multi-modal prompt tuning~\cite{zang2022unified}~\cite{khattak2023maple}~\cite{wang2023tuning}~\cite{shi2023logoprompt}~\cite{tu2023visual}~\cite{yao2023visual}~\cite{yao2021cpt}~\cite{xing2023dual}~\cite{cao2024aple} is an emerging task that facilitates the simultaneous learning of textual and visual prompts in VLMs.
Instead of independently optimizing uni-modal prompts, such a joint-tuning approach encourages interactions between the two modalities during training, leading to adaptable alignments.
Several studies proposed to enhance the generalization on downstream tasks by remaining knowledge inheritance and reduce the forgetting of the origin CLIP model ~\cite{yao2023visual}~\cite{yu2023task}~\cite{khattak2023self}~\cite{zang2024overcoming}.
Other methods~\cite{zhang2021tip}~\cite{zhang2023prompt} improve the 
% model's capability on 
downstream few-shot performance by adapting the output logits of VLMs.

\subsection{Adapters}
Adapters were first proposed as lightweight modules in the field of natural language processing~\cite{stickland2019bert}~\cite{houlsby2019parameter} for efficient model adaptation.
With the development of the vision-language model like CLIP~\cite{radford2021learning},
a large number of CLIP-based adapters have been proposed~\cite{gao2023clip}~\cite{zhang2021tip}~\cite{li2023graphadapter}~\cite{sung2022vl} to allow pre-trained models to better transfer to downstream tasks by inserting new learnable lightweight modules~\cite{chen2022vision}~\cite{li2021benchmarking}~\cite{li2022exploring}~\cite{chen2022adaptformer}.
These adapters enhance adaptation by either incorporating prior knowledge into the model or optimizing output representations of existing models.
In this paper,
we propose the "surgery" block, which is controlled by the feature shift to better alleviate the representation misalignment.
% is essentially an adapter influenced by the feature shift.

\section{Methodology}

\begin{figure}[!t]
	\centering
	\includegraphics[width=1.0\linewidth]{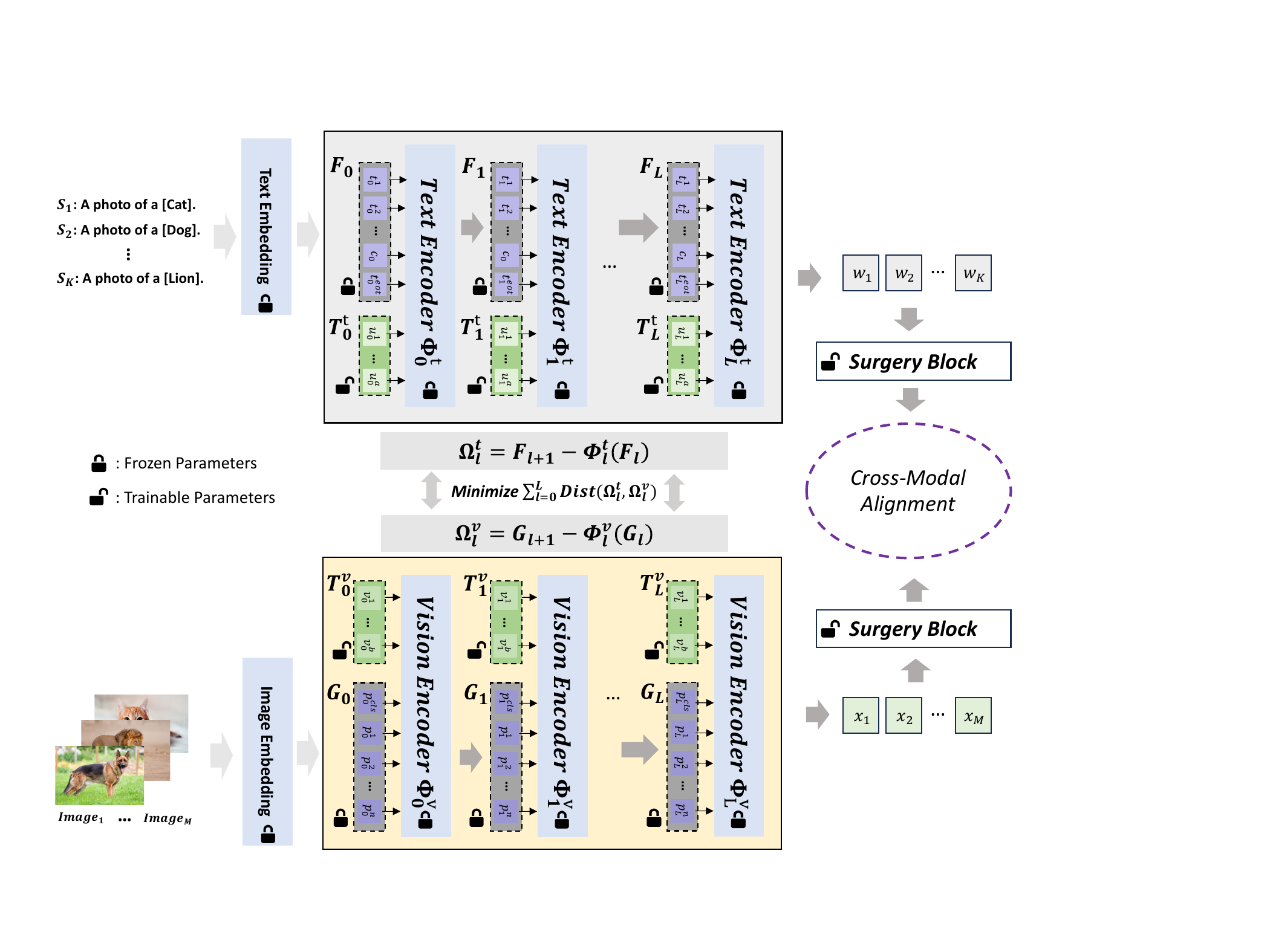}
	\caption{
 The overall workflow of our multi-modal prompt tuning.
 During fine-tuning,
 we fix the parameters of the encoder backbones unchanged.
 $K$ different textual descriptions are prompted to represent $K$ categories and are encoded by the text encoder into the embedding space.
 Similarly,
 $M$ images are encoded by the image encoder into the visual embedding space.
 The classification is carried out by measuring the similarity between visual and textual representations.
 In both vision and language encoders,
 multiple learnable prompts are equipped to interfere with the embeddings independently.
 % with that are combined with the image or text embedding.
 To establish connections between prompts from different modalities,
 we take feature shift as a bridge and synchronize the cross-modal representation update. 
 % an extra loss is used to align the feature shift caused by different modal prompts.
 % When there is a significant feature shift,
 In consideration of the risk of task-specific overfitting,
 the "surgery" block is applied to 
 % adapt outputs to downstream, which
 effectively penalize severe deviation of prompt-tuned features from their pre-trained counterparts,
 % aligned with its original pattern
 % and prevents deviating from
 preserving the valuable intrinsic knowledge.}
	\label{fig: framework}
\end{figure}

The proposed method aims to enhance the generalization capabilities of a pre-trained CLIP model for downstream tasks through multi-modal prompt tuning.
We start by introducing the process of contrastive language-image pre-training and task-specific fine-tuning, highlighting the intrinsic limitations of single-modal prompt tuning.
Subsequently,
as illustrated in Fig.~\ref{fig: framework},
we explain the mechanism of multi-modal prompt tuning.
Then,
the concept of feature shift is proposed,
followed by the definition of inter-modal discrepancy in feature shifts and its relationship with model generalization.
% across different modalities 
% Furthermore,
The cross-modal consistency is achieved by minimizing such inter-modal discrepancy.
Finally,
we present details about the implementation of "surgery" blocks,
especially around their control by feature shift to effectively address the deviated features.
% controllable represented by an adapter,
% This block dynamically corrects feature shifts in both the vision and language branches.

\subsection{Contrastive Vision Language Pre-training
	% Pre-trained Model
}

In the vision language pre-training framework,
positive samples are pairs of images and their corresponding texts, while mismatched images and texts serve as negative samples. Through the contrastive learning of positive and negative sample pairs, vision language models obtain strong generalization ability.
During the inference stages,
a manually designed prompt is incorporated into the textual component to create a zero-shot linear classifier.
This involves encoding class names
% relevant to
present in the target dataset into embeddings.
For instance, in the classification task, the "[CLASS]" token is first expanded into a prompt using a predefined template, such as "a photo of a [CLASS]."
Subsequently,
% the expanded sentence is defined as a prompt and 
the filled prompt is embedded by the text encoder 
% resulting in a set of embedding vectors
into the embedding space
% labeled 
as $F^{i}=\left\{\mathbf{t}^1, \mathbf{t}^2, \ldots, \mathbf{t}^m, \mathbf{c}^{i}, \mathbf{t}^{eot}\right\}, i \in [1, K]$, where $K$ is the total category number. Here $\mathbf{c}^{i}$ is the class token and $\mathbf{t}^{eot}$ is the end token of the sequence.
% At the same time,
Simultaneously, visual features are represented as $G=\left\{\mathbf{p}^{cls}, \mathbf{p}^1, \ldots, \mathbf{p}^n \right\}$.
We denote the language and vision encoders as $f$ and $g$ respectively. The latent text feature for $i_{th}$ category is defined as $\mathbf{w_i}=f(F^i)$. Similarly, the latent image feature is defined as $\mathbf{x}=g(G)$.
The class prediction probability $y$
% probability
can be computed as follows:
\begin{equation}
	p(y=i)=\frac{\exp \left(\operatorname{sim}\left(\mathbf{x}, \mathbf{w_i}\right) / \tau\right)}{\sum_{j=1}^K \exp \left(\operatorname{sim}\left(\mathbf{x}, \mathbf{w_j}\right) / \tau\right)},
\end{equation}
% \begin{equation}
% 	p(y=i)=\frac{\exp \left(\operatorname{sim}\left(g(G), f(F^i)\right) / \tau\right)}{\sum_{j=1}^K \exp \left(\operatorname{sim}\left(g(G), f(F^j)\right) / \tau\right)},
% \end{equation}
where $\operatorname{sim}(\cdot, \cdot)$ represents
% the computation of
cosine similarity,
and $\tau$ is the temperature coefficient.
% learned by CLIP.

\subsection{Multi-Modal Prompt Tuning}
Prompt tuning methods fine-tune the model by introducing learnable prompts. We denote
$\mathbf{T_{l}^{v}}=\left\{\mathbf{v}^1_{l}, \mathbf{v}^2_l, \ldots, \mathbf{v}^a_l\right\}$ and
$\mathbf{T_{l}^{t}}=\left\{\mathbf{u}^1_l, \mathbf{u}^2_l, \ldots, \mathbf{u}^b_l\right\}$
% denote the learnable text prompts, and
respectively as learnable vision and text prompts of the $l_{th}$ transformer block, where $a$ and $b$ represent the number of tokens that can be learned.
% denote the learnable vision prompt
The vision embedding is defined as the concatenation of learnable prompts and frozen image embeddings: $\left[\mathbf{T}_l^{v}, \mathbf{G}_l\right]$.
% denoted by
% $\left\{\mathbf{v}^1_l, \ldots, \mathbf{v}^b_l, \mathbf{p}^{cls}_l, \mathbf{p}^1_l, \ldots, \mathbf{p}^n_l \right\}$. 
The text embedding is defined as the concatenation of text prompts and class embeddings:$\left[\mathbf{T}_l^{t}, \mathbf{F}_l\right]$.
% $\left\{\mathbf{u}^1_l, \mathbf{u}^2_l, \ldots, \mathbf{u}^b_l, \mathbf{t}^1_l, \mathbf{t}^2_l, \ldots, \mathbf{c}^i_l, \mathbf{t}^{eot}_l\right\}$. 
% predictions can be calculated as follows:
% \begin{equation}
% 	p(y=i)=\frac{\exp \left(\operatorname{sim}\left(g(G, \{T^{v}_{l}\}_{l=1}^{L}), f(F^i, \{T^{t}_{l}\}_{l=1}^{L})\right) / \tau\right)}{\sum_{j=1}^K \exp \left(\operatorname{sim}\left(g(G, \{T^{v}_{l}\}_{l=1}^{L}), f(F^j, \{T^{t}_{l}\}_{l=1}^{L})\right) / \tau\right)},
% \end{equation}
% Since the input of the vision encoder and the text encoder have the same essential structure,
Out of simplicity, we present the formulation of vision prompt tuning as a demonstration and that of text prompt tuning can be derived similarly.
% we only take the input of the vision encoder as an example.
% Vanilla vision prompts are only introduced in shallow layers as follows:
The vanilla vision prompt tuning can be realized by interfering with visual features at each layer (the embeddings marked red are learnable):
\begin{equation}
	\begin{aligned}
		& {\left[\mathbf{T}_1^{v}, \mathbf{G}_1\right]=\Phi_0\left(\left[\textcolor{red}{\mathbf{T}_0^{v}}, \mathbf{G}_0\right]\right)}, \\
		& {\left[\mathbf{T}_l^{v}, \mathbf{G}_l\right]=\Phi_{l-1}\left(\left[ \mathbf{T}^{v}_{l-1}, \mathbf{G}_{l-1}\right]\right), l=2,3, \ldots, L,}
	\end{aligned}   
\end{equation}

 In this work, we follow the deep prompt tuning paradigm, where we introduce independent learnable prompts for each layer as follows:
 
\begin{equation}
	\left[\_, \mathbf{G}_l\right]=\Phi_{l-1}\left(\left[\textcolor{red}{\mathbf{T}^{v}_{l-1}}, \mathbf{G}_{l-1}\right]\right), l=1,2, \ldots, L .
\end{equation}

%Most of the previous work focused on the prompt of single modality, while our work follows \cite{khattak2023maple} and introduces learnable deep prompts in different modalities. 
%For deep multimodal prompt tuning, different modes are introduced into deep prompt to fine-tune the entire model. 

Therefore, the image feature $\mathbf{\Tilde{x}}$ and text feature $\mathbf{\Tilde{w_i}}$ of $i_{th}$ class can be obtained by $g(G, \{T^{v}_{l}\}_{l=1}^{L})$ and $f(F^i, \{T^{t}_{l}\}_{l=1}^{L})$, then the prediction can be calculated as follows:

\begin{equation}
	p(y=i)=\frac{\exp \left(\operatorname{sim}\left(\mathbf{\Tilde{x}}, \mathbf{\Tilde{w_i}}\right) \right)}{\sum_{j=1}^K \exp \left(\operatorname{sim}\left(\mathbf{\Tilde{x}}, \mathbf{\Tilde{w_j}}\right) \right)}.
\end{equation}

Given the one-hot label of the images $\hat{y}$, we use cross-entropy $\mathbf{\mathcal{L}^{ce}}$ as our final training loss:
\begin{equation}
    \mathbf{\mathcal{L}^{{ce}}}= -CrossEntropy\left(\hat{y}, y\right)
\end{equation}

\begin{figure}[!t] %
\centering
\includegraphics[width=1\linewidth]{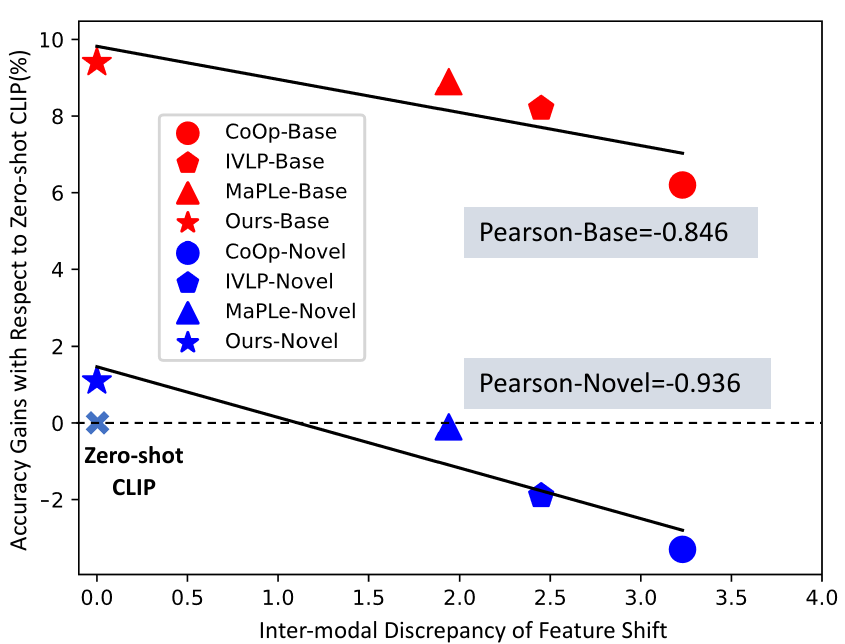}
\caption{
     The negatively associated relationship is observed between the inter-modal discrepancy of feature shift and the performance gains.
     Compared with the zero-shot CLIP,
     existing single-modal and multi-modal prompt tuning methods achieve superior and inferior performance respectively on base and novel categories.
     They "unintentionally" encourage the inter-modal discrepancy of feature shift during fine-tuning,
     % Those overfitting to base categories
     % are likely to experience cross-modal misalignment,
     consequently leading to a loss of generalization capabilities for downstream tasks.
 % The relationship between feature shift variation and corresponding performance. Here IVLP represents independent vision language prompt tuning. For the existing methods, the larger the feature shift variation, the worse the comprehensive performance of the model.
 }
\label{fig: fs_performance}
\end{figure}

\subsection{Feature Shift in Multi-Modal Prompt Tuning}
% \subsubsection{Feature Shift}
\subsubsection{Feature Shift.}
It is of great importance to impose proper constraints during prompt tuning for:
% Imposing constraints during the fine-tuning process of a model holds significant importance in
1) mitigating the issue of model collapse and
2) improving the efficiency of knowledge transfer for 
% the model's capacity to effectively transfer knowledge to
novel visual and textual concepts\cite{li2017learning,zhu2023prompt,li2023gradient,khattak2023self,yang2023pick}.
Various strategies have been put forth in the literature~\cite{kan2023knowledge,khattak2023self,zhu2023prompt},
with the primary aim of avoiding overfitting and model collapse.
Nonetheless,
none of these studies has paid enough attention to the underlying causes of model collapse in multi-modal systems. 
% In order to better understand the causes of model collapses in multi-modal systems,
Our assumption is that,
the pre-trained CLIP enjoys a high level of generalization over various downstream tasks thanks to its highly aligned cross-modal representations.
The degradation of such alignment,
which takes place in overfitting images or texts of specific domains or formats,
can be ascribed to the asynchronous, inconsistent updates of visual and textual features.
Such inter-modal discrepancy of features should be quantified for analysis of model generalization.
We first propose the concept of \textbf{feature shift}, which is used to estimate the variation of features generated by the vision-language model caused by the prompt tuning.
\textbf{Specifically, we define feature shift as the difference in feature representations of images or texts by transformer block with/without the introduction of learnable parameters.}
For the $l_{th}$ vision/text transformer block $\Phi_l^{v}/\Phi_l^{t}$,
given its corresponding inputs feature $G_{l}/F_{l}$ and learnable parameters $T_{l}^{v}/T_{l}^{t}$, the feature shift $\Omega_{l}^{v}$ and $\Omega_{l}^{t}$ are respectively defined as:
% \begin{equation}
%     \Omega_{l}^{v} = \Phi_l^{v}([T_{l}^{v}, G_{l}]) \ominus \Phi_l^{v}(G_{l})
% \end{equation}

\begin{equation}
	\begin{aligned}
		% &     \mathbf{\Omega_{l}^{v} = \Phi_l^{v}([T_{l}^{v},G_{l}])[:,...] - \Phi_l^{v}(G_{l})}, \\
		% &     \mathbf{\Omega_{l}^{t} = \Phi_l^{t}([T_{l}^{t},F_{l}])[:,...] - \ominus \Phi_l^{t}(F_{l})},
		   \mathbf{\Omega_{l}^{v}} &= \mathbf{G_{l+1} - \Phi_l^{v}(G_{l})},\ \ \ \ \mathbf{[\_, G_{l+1}]} = \mathbf{\Phi_l^{v}([T_{l}^{v},G_{l}])},\\
		\mathbf{\Omega_{l}^{t}} &= \mathbf{F_{l+1} - \Phi_l^{t}(F_{l})},\ \ \ \ \ \ \mathbf{[\_, F_{l+1}]} = \mathbf{\Phi_l^{t}([T_{l}^{t},F_{l}])}.\\
	\end{aligned}
\end{equation}
% \ominus
% \begin{equation}
%     \Omega_{l}^{t} = \Phi_l^{t}([T_{l}^{t}, F_{l}]) \ominus \Phi_l^{t}(F_{l})
% \end{equation}

% Since the inputs and outputs of the transformer are longer with the learnable prompts, we define $\ominus$ as a subtraction operation.
% It the model's outputs at the embedding positions corresponding to the image and text.

\subsubsection{Inter-modal Discrepancy of Feature Shift.}

We propose the feature shift as an explanatory tool to decipher such seesaw performance of prompt tuning methods (CoOp~\cite{zhou2022learning}, IVLP~\cite{rasheed2023fine,khattak2023self}, and MaPLe~\cite{khattak2023maple}) on base and novel classes.
By observing the inter-modal discrepancy of feature shift versus the performance gains (see Fig.~\ref{fig: fs_performance}), we can intuitively see the correlation that with larger feature shift variation, the model tends to perform worse.
Therefore, in Sec.~\ref{sec:featureshiftconsistency},
we propose feature shift consistency loss for cross-modal alignment.

\subsection{Feature Shift Consistency for Cross Modality Alignment}
\label{sec:featureshiftconsistency}
Our intuition
is that model misalignment is not only attributable to the deviation in either the visual or textual from the original CLIP embedding space.
Moreover, it stems from inconsistent feature shifts on both branches.
% It is worth noting that existing approaches predominantly 
% % center around the notion of
% \qyl{adopt the technique of}
% "learning without forgetting"\cite{zhou2023learning},
% % wherein the intention is to 
% \qyl{where} the fidelity of CLIP \qyl{is preserved as much as possible by minimizing the changes of parameters during fine-tuning.}
% % to its original pre-trained parameters and embeddings throughout the fine-tuning process.
% However,
% to retain both the original features and the post-tuning features, a frozen cache model is frequently introduced for storing images or text embeddings.
% This addition, although useful, comes at the expense of additional memory and computational resources\cite{zhang2023prompt}\cite{zhang2021tip}.
To enhance the cross-modal alignment, the discrepancies caused by the introduced prompts in different modality branches should be minimized. In practice, we try to minimize the distance of these feature shifts between different modalities in the feature space.
For the vision encoder and language encoder, given the feature shift on both vision and language branches as $\Omega_{l}^{v}$ and $\Omega_{l}^{t}$, the feature shift loss of the $l_{th}$ transformer block is defined as follows:
\begin{equation}
	\mathbf{L^{f s}_{l}}=\mathbf{Dist}\mathbf{\left(\Omega^{v}_{\textrm{l}}, \Omega^{t}_{\textrm{l}}\right)}
\end{equation}
where $\textbf{Dist}$ represents a measurement of the distance between feature variations of different modalities. Here we employ the mean square error between the matrix norm, and the final loss is defined as follows:
\begin{equation}
\begin{aligned}
     \mathbf{L^{f s}_{l}}=& \mathbf{MSE\left(Norm(\Omega_{l}^{\textrm{v}})-Norm(\Omega_{l}^{\textrm{t}})\right)} \\
        % &Norm(S^{l}) = ||T^T T||_{F} / ||T_{0}^{T}T_{0}||_{F}
\end{aligned}
\end{equation}
where $\mathbf{Norm}$ represents the Frobenius norm. For the feature shift of different transformer blocks, we use a hierarchical alignment strategy, and the full loss expression is defined as follows:
\begin{equation}
\label{norm_fs}
	\mathbf{\mathcal{L}^{fs}} = \mathbf{\sum_{l=1}^{L} L^{fs}_{l}}
\end{equation}
The overall loss function is the weighted summary of cross-entropy loss $\mathbf{\mathcal{L}^{ce}}$ and the feature shift loss $\mathbf{\mathcal{L}^{fs}}$ with weight coefficient $\lambda_{fs}$ as follows:
\begin{equation}
	\mathbf{\mathcal{L}_{total}} = \mathbf{\lambda_{fs} \mathcal{L}^{fs} + \mathcal{L}^{ce}}
\end{equation}

\subsection{Surgery Block with Feature Shift Guidance}

% After introducing the feature shift consistency constraint, we have more effective constraints on the features of different modalities.
One potential hacking way that bypasses the proposed cross-modal alignment constraint is to concurrently promote large feature shifts for both prompt-tuned visual and textual features,
where the layer-wise inter-modal discrepancy is estimated small but the overall accumulated deviation from pre-trained features can be huge.
The relative difference between feature shifts of two modalities can still be small (see Eq.~\ref{norm_fs}) if their Frobenius norm evolves with the same trend and at the same pace. 
% deactivates the proposed feature shift consistency
% The tokens used for classification in different modalities ($p^{cls}$ token on the vision side and $t^{eot}$ on the language side) will still appear misaligned when calculating similarity.
% This is due to the fact that feature shift loss solely imposes a constraint on the consistency of feature shift across various modalities, without directly controlling the feature shift within each individual modality.
Consequently,
the model is still prone to task-specific overfitting.
% feature shift loss can merely guarantee the synchronization of cross-modal feature variations.
In this case,
we propose the surgery block acting on both the two modalities in order to dynamically penalize cross-modal misalignment.
% when large feature shift is detected.
% adapt features and mitigate feature misalignment during prompt tuning on downstream data.
The operation of this surgery block is governed by the measured scale of feature shift,
meaning that a stronger correction effect is to be expected if a larger feature shift is detected.
% enabling dynamic weight adjustment in adaptation based on the current feature shift.
% Furthermore,the recently introduced surgery block can augment the feature representation and strengthen the model's capacity to represent new tasks, thereby ensuring the alignment pattern between vision and language remains consistent.
As depicted in Fig.~\ref{fig: framework},
% each output undergoes adjustment via the
the surgery block is implemented as an adapter as follows: 
%With respect to the implementation of such block,
%we resort to adapters since they are
% Adapters
%compact neural network modules that can be seamlessly integrated into the pre-trained model, enabling the adaptation to specific downstream tasks without extensive modifications to the entire pre-trained model.
% the purpose of adjusting the output features.
\begin{equation}
	\begin{aligned}
		\mathbf{\Tilde{x}}&=\mathbf{\alpha_{v}} * \mathbf{Surgery}(\mathbf{\Tilde{x}})+\mathbf{\Tilde{x}}, \\
		\mathbf{\Tilde{w_i}}&=\mathbf{\alpha_{t}} * \mathbf{Surgery}(\mathbf{\Tilde{w_i}})+\mathbf{\Tilde{w_i}}, \\
		\mathbf{\alpha_{v}} &= \mathbf{\gamma \sum_{l=1}^{L}Norm(\Omega^{v}_{l})}, \\
		\mathbf{\alpha_{t}} &= \mathbf{\beta \sum_{l=1}^{L}Norm(\Omega^{t}_{l})},\\	\mathbf{Surgery(\Tilde{x})}&=\mathbf{ReLU}\left(\boldsymbol{LN}\left(\mathbf{\Tilde{x}}\right) \cdot \boldsymbol{W}_{\text {up}}\right) \cdot \boldsymbol{W}_{\text {down}},
	\end{aligned}
\end{equation}
where $\gamma$, $\beta$ are hyper-parameters.
$\boldsymbol{W}_{\text {up}}$ and $\boldsymbol{W}_{\text {down}}$ represent up-scale and down-scale linear mappings, respectively. $\boldsymbol{LN}$ represents the layer normalization.
% linear mapping to raise and lower dimensions.
We dynamically update $\alpha_{v}$ and $\alpha_{t}$ to control the surgery during training.
%, the feature shift information is fed to the surgery block so as to eliminate the misalignment problem. 

\section{Experiments}

\subsection{Datasets and Implementation Details}

\noindent\textbf{Datasets.} Following previous prompt tuning studies, we validate our method on 11 few-shot classification datasets, including ImageNet\cite{deng2009imagenet}, StanfordCars\cite{krause20133d}, UCF101\cite{soomro2012ucf101}, Caltech101\cite{fei2004learning}, Flowers102\cite{nilsback2008automated}, SUN397\cite{xiao2010sun}, DTD\cite{cimpoi2014describing}, EuroSAT\cite{helber2019eurosat}, FGVCAircraft\cite{maji2013fine}, OxfordPets\cite{parkhi2012cats} and Food101\cite{bossard2014food}. OxfordPets, Food101, StanfordCars, Flowers102, and FGVCAircraft belong to fine-grained classification tasks, EuroSAT is for remote sensing classification, and DTD is the dataset of texture classification. The partitioning of all datasets follows \cite{zhou2022conditional}\cite{zhou2022learning}. 

\noindent\textbf{Implementation Details}. We adopt a few-shot training strategy in all experiments at 16 shots which are randomly sampled for each class. We apply our multi-modal prompt tuning method on a pre-trained CLIP model with ViT-B/16 \cite{dosovitskiy2020image} as the image encoder. Each model is trained with a batch size of 4 and a learning rate of 0.0035 via SGD optimizer on a single NVIDIA RTX3090 GPU. For cross-dataset evaluation and cross-domain evaluation, each model is trained for 4 epochs due to computing resources constraint while in base-to-novel evaluation the training epochs is 10. Besides, we report the results over 3 different random seeds to make results reliable. For a fair comparison, the text description of each class 
% in the first layer
is simply initialized as
% by the embedding of
"a photo of a [CLASS]" for each method.

\subsection{Base-to-Novel Evaluation}

We first follow \cite{khattak2023maple} to evaluate our method under the base-to-novel setting, where we evenly divide datasets into two groups: the base classes and the novel classes. The model undergoes training on the base classes and is subsequently assessed on its performance with respect to both the base and unseen novel classes. This benchmark serves as a means to gauge the model's generalization capability.

\begin{table}[htbp]
  \centering
  \caption{Base-to-Novel evaluation. We uniformly chose the same epoch number (10 epochs) for the fairness of comparison. We reproduce the results of prompt tuning methods and build our method on the basis of MaPLe and PromptSRC methods, with $\mathbf{RESTORE_{m}}$ implemented based on MaPLe and $\mathbf{RESTORE_{p}}$ based on PromptSRC. HM is the harmonic mean of classification accuracy on base class and novel class.}
  \label{tab:base2new}
  \begin{subtable}{0.3\textwidth}
    \centering
    \caption{Average on 11 datasets}
    \scalebox{0.6}{
    \begin{tabular}{lcc|c}
        \hline \textbf{Method} & \textbf{Base} & \textbf{Novel} & \textbf{HM} \\
    \hline 
    \textit{\textbf{Zero-shot}} \\
    CLIP & 69.34 & 74.22 & 71.70 \\
        \hline
        \textit{\textbf{Prompt Tuning}} \\
    CoOp & 81.84 & 70.94 & 76.00 \\
    Co-CoOp & 79.38 & 73.11 & 76.11 \\
    \hline 
    
    MaPLe & 83.47 & 74.09 & 78.50 \\
        \rowcolor{gray!20}
    $RESTORE_{m}$ & 83.93 & 75.60 & \textbf{79.55}\textbf{(+1.1)} \\
    \hdashline
    PromptSRC & 83.14 & 75.11 & 78.92 \\

    \rowcolor{gray!20}
    $RESTORE_{p}$ & 83.44 & 76.35 & \textbf{79.74}\textbf{(+0.8)} \\
    \hline
    \end{tabular}
    }
  \end{subtable}
  \hfill
  \begin{subtable}{0.3\textwidth}
    \centering
    \caption{ImageNet}
        \scalebox{0.6}{
        \begin{tabular}{lcc|c}
        \hline \textbf{Method} & \textbf{Base} & \textbf{Novel} & \textbf{HM} \\
        \hline
        \textit{\textbf{Zero-shot}} \\
        CLIP & 72.43 & 68.14 & 70.22 \\
        \hline
        \textit{\textbf{Prompt Tuning}} \\
        CoOp & 76.67 & 69.73 & 73.03 \\
        Co-CoOp & 75.99 & 70.77 & 73.28 \\
        \hline MaPLe & 77.11 & 69.49 & 73.10 \\
        \rowcolor{gray!20}
        $RESTORE_{m}$ & 77.42 & 69.83 & \textbf{73.43}\textbf{(+0.3)} \\
            \hdashline
            PromptSRC & 77.30 & 70.66 & 73.83 \\
        \rowcolor{gray!20}
        $RESTORE_{p}$ & 77.39 & 70.88 & \textbf{73.99}\textbf{(+0.2)} \\
        \hline
        \end{tabular}
        }
  \end{subtable}
  \hfill
  \begin{subtable}{0.3\textwidth}
    \centering
    \caption{Caltech101}
        \scalebox{0.6}{
        \begin{tabular}{lcc|c}
        \hline \textbf{Method} & \textbf{Base} & \textbf{Novel} & \textbf{HM} \\
        \hline 
            \textit{\textbf{Zero-shot}} \\
        CLIP & 96.84 & 94.00 & 95.40 \\
        \hline
        \textit{\textbf{Prompt Tuning}} \\
        CoOp & 97.94 & 94.32 & 96.09 \\
        Co-CoOp & 97.72 & 93.08 & 95.34 \\
        \hline MaPLe & 98.41 & 95.05 & 96.70 \\
        \rowcolor{gray!20}
        $RESTORE_{m}$ & 98.30 & 95.61 & \textbf{96.94}\textbf{(+0.2)} \\
            \hdashline
            PromptSRC & 97.83 & 94.25 & 96.00 \\
        \rowcolor{gray!20}
        $RESTORE_{p}$ & 97.84 & 94.84 & \textbf{96.32}(\textbf{+0.3)} \\
        \hline
        \end{tabular}
        }
  \end{subtable}
  \\
  \vspace{0.5cm}
  \begin{subtable}{0.3\textwidth}
    \centering
    \caption{OxfordPets}
        \scalebox{0.6}{
    \begin{tabular}{lcc|c}
        \hline \textbf{Method} & \textbf{Base} & \textbf{Novel} & \textbf{HM} \\
    \hline 
        \textit{\textbf{Zero-shot}} \\
    CLIP & 91.17 & 97.26 & 94.12 \\
        \hline
        \textit{\textbf{Prompt Tuning}} \\
    CoOp & 95.55 & 97.43 & 96.48 \\
    Co-CoOp & 94.84 & 97.60 & 96.20 \\
    \hline MaPLe & 95.16 & 97.78 & 96.45 \\
            \rowcolor{gray!20}
        $RESTORE_{m}$ & 95.59 & 98.03 & \textbf{96.79}\textbf{(+0.3)} \\
            \hdashline
        PromptSRC & 95.50 & 97.16 & 96.32 \\
        \rowcolor{gray!20}
        $RESTORE_{p}$ & 95.56 & 98.04 & \textbf{96.78}\textbf{(+0.4)} \\
    \hline
    \end{tabular}
        }
  \end{subtable}
  \hfill
  \begin{subtable}{0.3\textwidth}
    \centering
    \caption{StanfordCars}
        \scalebox{0.6}{
        \begin{tabular}{lcc|c}
        \hline \textbf{Method} & \textbf{Base} & \textbf{Novel} & \textbf{HM} \\
        \hline 
            \textit{\textbf{Zero-shot}} \\
        CLIP & 63.37 & 74.89 & 68.65 \\
        \hline
        \textit{\textbf{Prompt Tuning}} \\
        CoOp & 74.49 & 70.87 & 72.63 \\
        Co-CoOp & 69.87 & 74.69 & 72.20 \\
        \hline MaPLe & 76.22 & 73.08 & 74.62 \\
                \rowcolor{gray!20}
        $RESTORE_{m}$ & 78.10 & 73.29 & \textbf{75.61}\textbf{(+1.0)} \\
            \hdashline
            PromptSRC & 76.04 & 75.41 & 75.72 \\
        \rowcolor{gray!20}
        $RESTORE_{p}$ & 76.61 & 76.01 & \textbf{76.31}\textbf{(+0.6)} \\
        \hline
        \end{tabular}
        }
  \end{subtable}
  \hfill
  \begin{subtable}{0.3\textwidth}
    \centering
    \caption{Flowers102}
        \scalebox{0.6}{
        \begin{tabular}{lcc|c}
        \hline \textbf{Method} & \textbf{Base} & \textbf{Novel} & \textbf{HM} \\
        \hline 
            \textit{\textbf{Zero-shot}} \\
        CLIP & 72.08 & 77.80 & 74.83 \\
        \hline
        \textit{\textbf{Prompt Tuning}} \\
        CoOp & 96.96 & 72.41 & 82.89 \\
        Co-CoOp & 93.00 & 72.84 & 81.69 \\
        \hline MaPLe & 96.99 & 73.59 & 83.69 \\
                \rowcolor{gray!20}
        $RESTORE_{m}$ & 97.66 & 74.30 & \textbf{84.39}\textbf{(+0.7)} \\
            \hdashline
            PromptSRC & 97.69 & 76.53 & 85.82 \\
        \rowcolor{gray!20}
        $RESTORE_{p}$ & 97.63 & 76.74 & \textbf{85.93}\textbf{(+0.1)} \\
        \hline
        \end{tabular}
        }
  \end{subtable}
  \\
  \vspace{0.5cm}
  \begin{subtable}{0.3\textwidth}
    \centering
    \caption{Food101}
        \scalebox{0.6}{
        \begin{tabular}{lcc|c}
        \hline \textbf{Method} & \textbf{Base} & \textbf{Novel} & \textbf{HM} \\
        \hline 
      \textit{\textbf{Zero-shot}} \\
        CLIP & 90.10 & 91.22 & 90.66 \\
        \hline
        \textit{\textbf{Prompt Tuning}} \\
        CoOp & 90.32 & 91.23 & 90.77 \\
        Co-CoOp & 90.68 & 91.71 & 91.19 \\
        \hline MaPLe & 90.07 & 91.54 & 90.80 \\
                \rowcolor{gray!20}
        $RESTORE_{m}$ & 90.46 & 91.74 & \textbf{91.10}\textbf{(+0.3)} \\
            \hdashline
            PromptSRC & 90.69 & 91.74 & 91.21 \\
        \rowcolor{gray!20}
        $RESTORE_{p}$ & 90.82 & 91.75 & \textbf{91.28}\textbf{(+0.1)} \\
        \hline
        \end{tabular}
        }
  \end{subtable}
  \hfill
  \begin{subtable}{0.3\textwidth}
    \centering
    \caption{FGVCAircraft}
        \scalebox{0.6}{
        \begin{tabular}{lcc|c}
        \hline \textbf{Method} & \textbf{Base} & \textbf{Novel} & \textbf{HM} \\
        \hline 
            \textit{\textbf{Zero-shot}} \\
        CLIP & 72.43 & 68.14 & 70.22 \\
        \hline
        \textit{\textbf{Prompt Tuning}} \\
        CoOp & 37.98 & 32.35 & 34.93 \\
        Co-CoOp & 36.09 & 34.21 & 35.12 \\
        \hline MaPLe & 39.88 & 34.03 & 36.70 \\
                \rowcolor{gray!20}
        $RESTORE_{m}$ & 40.62 & 36.37 & \textbf{38.36}\textbf{(+1.7)} \\
            \hdashline
            PromptSRC & 37.66 & 28.27 & 31.27 \\
        \rowcolor{gray!20}
        $RESTORE_{p}$ & 39.58 & 36.85 & \textbf{38.17}\textbf{(+6.9)} \\
        \hline
        \end{tabular}
        }
  \end{subtable}
  \hfill
  \begin{subtable}{0.3\textwidth}
    \centering
    \caption{SUN397}
        \scalebox{0.6}{
        \begin{tabular}{lcc|c}
        \hline \textbf{Method} & \textbf{Base} & \textbf{Novel} & \textbf{HM} \\
        \hline 
            \textit{\textbf{Zero-shot}} \\
        CLIP & 69.36 & 75.35 & 72.23 \\
        \hline
        \textit{\textbf{Prompt Tuning}} \\
        CoOp & 80.83 & 75.30 & 77.96 \\
        Co-CoOp & 79.27& 76.84 & 78.03 \\
        \hline MaPLe & 81.68 & 77.58 & 79.57 \\
                \rowcolor{gray!20}
        $RESTORE_{m}$ & 81.97 & 77.31 & \textbf{79.57}\textbf{(+0.0)} \\
            \hdashline
            PromptSRC & 82.44 & 78.75 & 80.55 \\
        \rowcolor{gray!20}
        $RESTORE_{p}$ & 82.52 & 78.88 & \textbf{80.66}\textbf{(+0.1)} \\
        \hline
        \end{tabular}
        }
  \end{subtable}
  \\
  \vspace{0.5cm}
  \begin{subtable}{0.3\textwidth}
    \centering
    \caption{DTD}
        \scalebox{0.6}{
        \begin{tabular}{lcc|c}
        \hline \textbf{Method} & \textbf{Base} & \textbf{Novel} & \textbf{HM} \\
        \hline 
            \textit{\textbf{Zero-shot}} \\
        CLIP & 53.24 & 59.90 & 56.37 \\
        \hline
        \textit{\textbf{Prompt Tuning}} \\
        CoOp & 78.90 & 48.35 & 59.78 \\
        Co-CoOp & 73.80 & 56.04 & 63.70 \\
        \hline MaPLe & 81.36 & 59.02 & 68.40 \\
                    \rowcolor{gray!20}
        $RESTORE_{m}$ & 82.79 & 61.83 & \textbf{70.79}\textbf{(+2.4)} \\
            \hdashline
            PromptSRC & 82.67 & 60.99 & 70.18 \\
        \rowcolor{gray!20}
        $RESTORE_{p}$ & 82.69 & 61.64 & \textbf{70.63}\textbf{(+0.4)} \\
        \hline
        \end{tabular}
        }
  \end{subtable}
  \hfill
  \begin{subtable}{0.3\textwidth}
    \centering
    \caption{EuroSAT}
        \scalebox{0.6}{
        \begin{tabular}{lcc|c}
        \hline \textbf{Method} & \textbf{Base} & \textbf{Novel} & \textbf{HM} \\
        \hline 
            \textit{\textbf{Zero-shot}} \\
        CLIP & 56.48 & 64.05 & 60.03 \\
        \hline
        \textit{\textbf{Prompt Tuning}} \\
        CoOp & 86.94 & 56.14 & 67.91 \\
        Co-CoOp & 80.94 & 63.98 & 71.29 \\
        \hline MaPLe & 95.73 & 67.59 & 79.07 \\
                    \rowcolor{gray!20}
        $RESTORE_{m}$ & 94.71 & 75.97 & \textbf{84.04}\textbf{(+5.0)} \\
            \hdashline
            PromptSRC & 90.69 & 73.86 & 81.37 \\
        \rowcolor{gray!20}
        $RESTORE_{p}$ & 91.50 & 74.45 & \textbf{81.79}\textbf{(+0.4)} \\
        \hline
        \end{tabular}
        }
  \end{subtable}
  \hfill
  \begin{subtable}{0.3\textwidth}
    \centering
    \caption{UCF101}
        \scalebox{0.6}{
        \begin{tabular}{lcc|c}
        \hline \textbf{Method} & \textbf{Base} & \textbf{Novel} & \textbf{HM} \\
        \hline 
            \textit{\textbf{Zero-shot}} \\
        CLIP & 70.53 & 77.50 & 73.85 \\
        \hline
        \textit{\textbf{Prompt Tuning}} \\
        CoOp & 83.63 & 72.20 & 77.50 \\
        Co-CoOp & 80.94 & 72.45 & 76.46 \\
        \hline MaPLe & 85.51 & 76.24 & 80.60 \\
                    \rowcolor{gray!20}
        $RESTORE_{m}$ & 85.66 & 77.37 & \textbf{81.31}\textbf{(+0.7)} \\
            \hdashline
            PromptSRC & 86.02 & 78.58 & 82.13 \\
        \rowcolor{gray!20}
        $RESTORE_{p}$ & 85.95 & 79.77 & \textbf{82.75}\textbf{(+0.6)} \\
        \hline
        \end{tabular}
        }
  \end{subtable}
\end{table}

Table \ref{tab:base2new} presents the comparison between the proposed method and baselines including Zero-shot CLIP,
% DPT~\cite{xing2023dual}, KgCoOp~\cite{yao2023visual}, ProGrad~\cite{zhu2023prompt}, 
CoOp~\cite{zhou2022learning}, Co-CoOp~\cite{zhou2022conditional}, MaPLe~\cite{khattak2023maple}, PromptSRC\cite{khattak2023self} on 11 datasets. 
% MaPLe\cite{khattak2023maple} achieves improved results by combining the vision branch and the language prompt using an MLP, but this combination lacks an explicit constraint. PromptSRC\cite{khattak2023self} introduces frozen CLIP to regularize the network while neglecting the alignment of prompts from different modalities.
Our approach directly imposes stronger alignment constraints on different modality branches hierarchically, and introduces the surgery for the output feature to alleviate the misalignment problem. This not only improves the performance on the base class but also improves the performance on the unseen class clearly.

\subsection{Cross Dataset/Domain Evaluation}
To further validate the generalization ability of our method, we tuned our model on ImageNet and tested on other datasets. Under the cross-dataset setting, each model is tested on the other ten target datasets, while under the cross-domain setting, we assess the performance on several variants of ImageNet: ImageNetV2\cite{recht2019imagenet}, ImageNet-Sketch\cite{wang2019learning}, ImageNet-A\cite{hendrycks2021natural}, and ImageNet-R\cite{hendrycks2021many}. As depicted in Table \ref{tab::cross_dataset} and \ref{tab::cross_domain}, our method surpasses other approaches on the majority of the datasets, demonstrating its ability to maintain an exceptional level of generalization even in situations with significant domain gaps.

\begin{table*}[!t]
	\centering
	\caption{Cross dataset evaluation. The model is trained on ImageNet and tested on 10 unseen target datasets. This experiment mainly evaluates the generalization ability of the model between different datasets.}
	\label{tab::cross_dataset}
	\scalebox{0.85}{
            \begin{tabular}{c|c|ccc>{\columncolor{gray!15}}cc>{\columncolor{gray!15}}c}
            \hline
                                     \textbf{Domain} & \textbf{Dataset}         & \textbf{Coop}    & \textbf{Co-CoOp}  & \textbf{MaPLe}   & $\mathbf{RESTORE_{m}}$ &  \textbf{PromptSRC}   & $\mathbf{RESTORE_{p}}$ \\
                                     \hline
            \textbf{Source}                   & ImageNet        & 70.61\% & 70.45\% & 70.78\% & 71.43\% \textbf{(+0.7)}  & 71.34\% & 71.55\%\textbf{(+0.2)}       \\
            \hline
            \multirow{10}{*}{\textbf{Target}} & Caltech101      & 93.87\% & 93.27\% & 93.35\% & 93.39\%   &  93.26\%& 93.69\%        \\
                                     & OxfordPets    & 90.13\% & 90.68\% & 89.92\% & 90.35\%   & 90.21\% & 90.95\%      \\
                                     & StanfordCars  & 65.70\% & 65.71\% & 65.91\% & 66.46\%   & 65.70\% & 66.96\%     \\
                                     & Flowers102 & 71.21\% & 72.35\% & 71.01\% &   71.90\% & 70.25\% & 71.95\%       \\
                                     & Food101         & 86.13\% & 86.10\% & 86.40\% &  86.55\%  & 86.15\% & 87.43\%        \\
                                     & FGVCaircraft  & 22.47\% & 22.11\% & 24.81\% &  24.97\%   & 24.14\% & 23.93\%        \\
                                     & SUN397          & 66.73\% & 66.71\% & 66.23\% & 67.69\%  &  67.09\% & 68.29\%        \\
                                     & DTD             & 44.64\% & 47.04\% & 45.27\% & 47.10\%   &  46.44\%& 45.21\%        \\
                                     & EuroSAT       & 44.56\% & 44.28\% & 42.80\% & 45.00\%    & 44.34\% & 49.34\%        \\
                                     & UCF101          & 67.38\% & 67.33\% & 66.69\% & 67.78\%   &  68.21\% & 68.21\%      \\
                                     \cdashline{2-8}
                                     & Average         & 65.28\% & 65.56\% & 65.24\% & 66.12\%\textbf{(+0.9)} & 65.58\% & 66.60\%\textbf{(+1.0)}     \\
                                     \hline
            \end{tabular}
	}
\end{table*}

\begin{table*}[!t]
	\centering
	\caption{Cross domain validation on 4 datasets. Models are trained on ImageNet and tested on several datasets with certain domain gaps.}
	\label{tab::cross_domain}
	\scalebox{0.85}{
        \begin{tabular}{c|c|ccc>{\columncolor{gray!15}}cc>{\columncolor{gray!15}}c}
        \hline
                                \textbf{Domain} & \textbf{Dataset}    & \textbf{Coop}    & \textbf{Co-CoOp}  & \textbf{MaPLe}   & $\mathbf{RESTORE_{m}}$ &  \textbf{PromptSRC} & $\mathbf{RESTORE_{p}}$  \\
                                \hline
        \textbf{Source}                  & ImageNet   & 70.61\% & 70.45\% & 70.78\% & 71.43\%\textbf{(+0.7)}   & 71.34\% & 71.55\%\textbf{(+0.2)} \\
        \hline
        \multirow{4}{*}{\textbf{Target}} & ImageNetV2 & 63.94\% & 63.83\% & 64.14\% & 64.16\%   & 64.32\% & 64.67\% \\
                                & ImageNet-S & 49.05\% & 48.81\% & 48.82\% & 49.11\%   & 49.37\% & 49.93\% \\
                                & ImageNet-A & 51.03\% & 50.88\% & 50.16\% & 50.51\%  & 51.23\% & 51.45\% \\
                                & ImageNet-R & 77.00\% & 76.98\% & 76.75\% & 77.17\%   & 77.88\% & 77.64\% \\
                                \cdashline{2-8}
                                & Average    & 60.26\% & 60.33\% & 59.97\% & 60.24\%\textbf{(+0.3)}  & 60.70\% & 60.92\%\textbf{(+0.2)} \\
                                \hline
        \end{tabular}
	}
\end{table*}

\begin{figure*}[!t]
	\centering
	\includegraphics[width=1.0\linewidth]{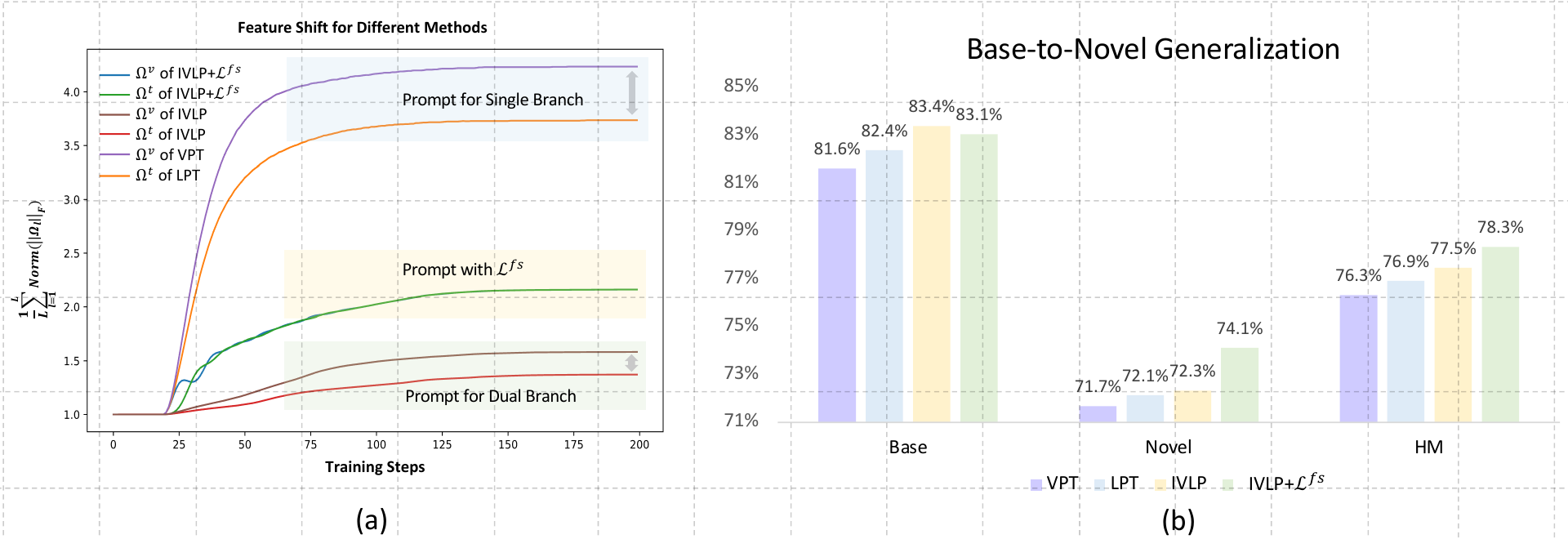}
	\caption{Average feature shift and the according performance for different methods. IVLP, VPT, LPT, and IVLP+$\mathcal{L}^{fs}$ represent independent vision-language prompt tuning, vision prompt tuning, language prompt tuning,  and IVLP with feature shift loss. The introduction of prompt tuning in a single branch causes severe feature shifts, leading to final feature misalignment and degradation of performance. However, the introduction of our feature shift loss can reduce such kind of modality misalignment, therefore causing superior performance.}
	\label{fig:: feat_shift}
\end{figure*}
\subsection{Ablation}
In this section we explore the effectiveness of the various modules that have been proposed modules.

\noindent\textbf{Effectiveness of Feature Shift Loss.} We conducted an investigation into the impact of the feature shift loss. By adjusting the coefficients of the loss, we compared the performance changes of the base and novel classes on different datasets. Table \ref{tab::ablation_fs} presents that as the coefficient increases, the performance on the base class decreases while the performance on the novel class increases. This observation demonstrates that the feature shift loss, which aligns modalities, effectively maintains and improves the alignment ability of the vision-language model. Taking into account the results for both the base and novel classes, we set the coefficient $\mathbf{\lambda_{fs}}$ to 1.% In Fig. \ref{fig:ivlp_epoch} we demonstrate that with the increase of the training epoch, the performance of independent vision-language prompt(IVLP) in the base category will continue to improve, while the performance in the novel category will continue to decline. After adding fs loss, the performance degradation on the base class becomes smaller, which proves that our method alleviates feature misalignment.

\begin{table}[htbp]
  \begin{minipage}[b]{0.45\linewidth}
    \centering
    \scalebox{1.0}{
		\begin{tabular}{m{0.8cm}|ccc}
			\hline
			              $\mathbf{\lambda_{fs}}$                 & \textbf{Base} & \textbf{Novel} &  \textbf{HM} \\
			\hline
			0   &         83.80\%         &        75.24\%         &            79.29\%                \\
			% 0.1 &        83.81\%          &         74.52\%        &      78.79\%      \\
			0.5   &        83.64\%          &        75.79\%         &     79.52\%       \\
			1   &         83.44\%         &        76.35\%         &      \textbf{79.74\%}      \\
			2   &         83.12\%         &        76.45\%         &      79.64\%     \\
			\hline
		\end{tabular}
  }
    \caption{Ablation of the feature shift loss on base-to-novel transfer reveals its significant contribution to the novel class. To strike a balance between the performance on both the base and novel classes, we set the $\lambda_{fs}$ to 1.}
    \label{tab::ablation_fs}
  \end{minipage}
  \hfill
  \begin{minipage}[b]{0.5\linewidth}
    \centering
     	\scalebox{0.99}{
	\begin{tabular}{c|ccc}
		\hline
		\textbf{Methods}                             & \textbf{Base} & \textbf{Novel} &  \textbf{HM} \\
		\hline
		\textbf{Baseline}   &         83.10\%         &       76.00\%          &           79.39\%               \\
		\textbf{+Surgery}   &         83.20\%         &       76.09\%          &          79.48\%                 \\
		% FS Fixed &         83.80\%         &      75.01\%           &       78.96\%    \\
		\textbf{+FS-Guide} &       83.44\%           &     76.35\%           &      \textbf{79.74\%}      \\
		
		\hline
	\end{tabular}
        }
    \caption{Ablation study on controllable surgery block. We compare the difference between baseline (only with feature shift consistency loss), using a fixed surgery coefficient(+Surgery), and dynamically adjusting the coefficients via feature shift (+FS-Guided).}
    \label{tab:: ablation_adapter}
  \end{minipage}
\end{table}

% \begin{table}[]
% 	\caption{Ablation study on feature shift loss on base-to-novel transfer. The feature shift loss mainly contributes to the new class performance rather than the base class.}
% 	\label{tab::ablation_fs}
% 	\scalebox{1}{
% 		\begin{tabular}{c|ccc}
% 			\hline
% 			            $\lambda_{fs}$                 & Base Accuracy & New Accuracy &  HM \\
% 			\hline
% 			0   &         83.80\%         &        74.12\%         &            78.61\%                \\
% 			0.1 &        83.81\%          &         74.52\%        &      78.79\%      \\
% 			0.5   &        83.64\%          &        74.79\%         &     78.91\%       \\
% 			1   &         \textbf{83.90\% }        &        75.20\%         &      \textbf{79.02\%}      \\
% 			2   &         83.42\%         &        \textbf{75.59\%}         &      78.96\%     \\
% 			\hline
% 		\end{tabular}
% 	}
% \end{table}

\noindent\textbf{Effectiveness of FS-Guided Surgery Block.} As shown in Table \ref{tab:: ablation_adapter}, our surgery block offers greater advantages for the new class rather than the base class. This can be attributed to the adapter's precise adaptation of the final embedding, leading to improved generalization. Additionally, we compared adapters with fixed coefficients to those with controllable coefficients. Results indicate that adapters with controllable coefficients outperform their counterparts, which is due to the fact that the observed shifts in features necessitate correction, and the controlled coefficients are more effective in accurately rectifying these features.

\begin{figure}[!t]
	\centering
	\includegraphics[width=1\linewidth]{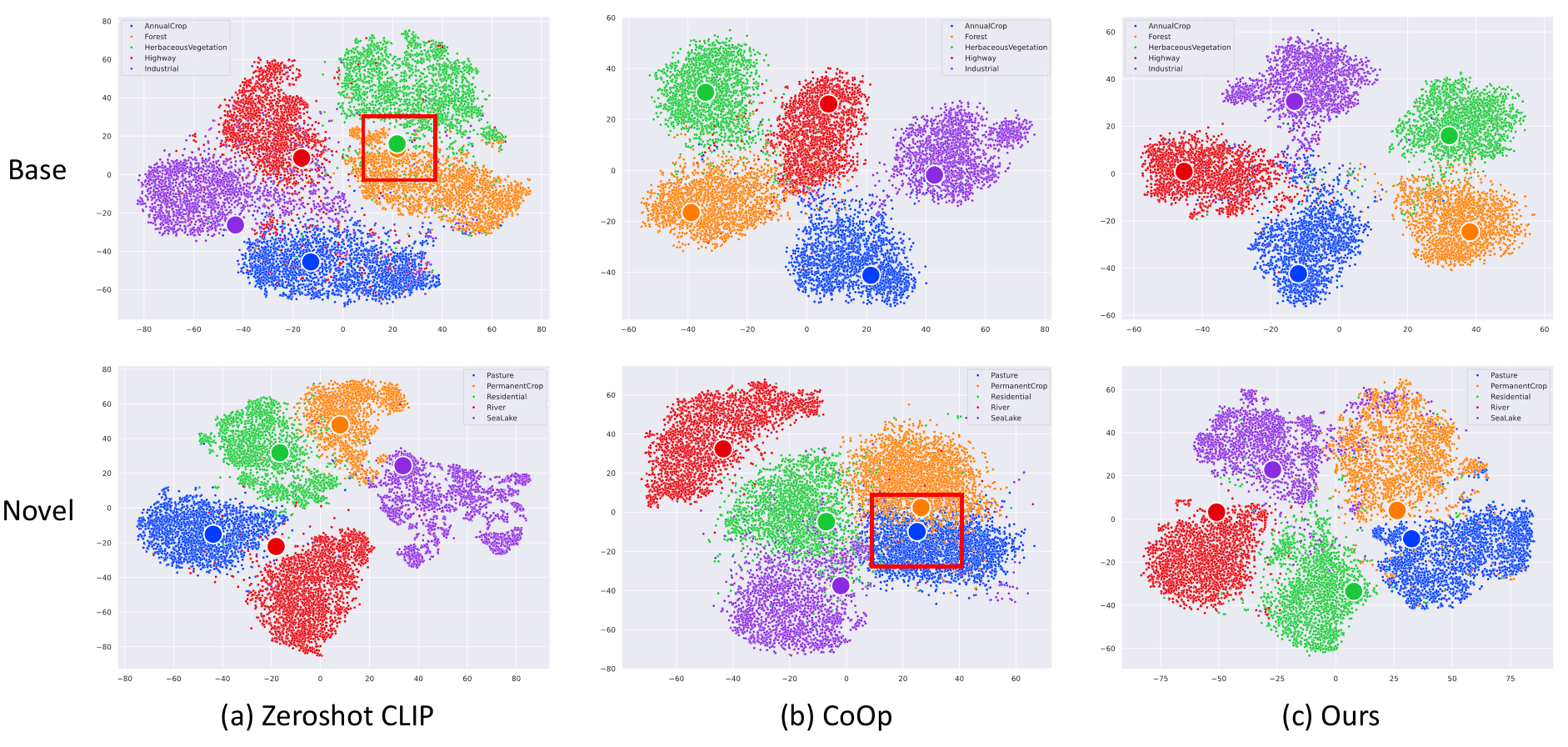}
	\caption{T-SNE visualization of features on base and novel classes after training with CoOp, our method and zero-shot CLIP. Different colored dots in the figure represent different categories, with smaller dots representing t-SNE visualizations of image features and larger dots representing t-SNE visualizations of text features. The zero-shot CLIP performs poorly on the base class, while the fine-tuned CoOp performs poorly on the new category (see the red box in the figure). However, after introducing cross-modal constraints and an adapter to alleviate feature collapse, our method performs very well on both the base class and the novel class.
}
\label{fig:: tsne visualization}
\end{figure}

\section{Visualization and Analysis}
\noindent\textbf{Feature Shift Analysis.}
The relationship between feature shift and performance is investigated in this part. We add prompts on the vision branch, prompts on the language branch, and unrelated prompts on both the visual and language branch to the CLIP model during the fine-tuning process, and compare the feature shift as well as the final model performance. As shown in Fig.~\ref{fig:: feat_shift} (a), we find that the feature shift generated by using only vision prompt or language prompt is larger. This is because there are fewer learnable parameters, so the features of a single modality need to change more to adapt to downstream tasks, which can easily cause overfitting. Due to the introduction of more parameters, the value of the feature shift of IVLP is relatively small, but its variation still exists, which poses a challenge to the alignment ability of models. After the addition of feature shift loss, although the shift of the model will increase, its variation will decrease and its generalization ability will be stronger. From Fig.~\ref{fig:: feat_shift} (b) it is evident that the generalization performance of the model is subpar when using solely vision-side prompts or text-side prompts compared to when both types are combined. After adding $\mathcal{L}^{fs}$, we find that the performance of the model is further improved, which proves the effectiveness of our feature shift loss.

\noindent\textbf{T-SNE Visualization.} We conducted t-SNE visualization (refer to Fig. \ref{fig:: tsne visualization}) on features generated by various methods. Our observations revealed that the zero-shot CLIP model displayed subpar performance in the base class, while the CoOp model (vanilla prompt tuning) exhibited limited generalization in the novel class. Conversely, our proposed approach outperformed both the base and novel classes, showcasing superior results.

% \begin{table}[]
% 	\caption{Ablation study on controllable adapter. We compare the difference between randomly fixing the adapter coefficients(Fixed), using FS as fixed coefficients(FS Fixed), and dynamically adjusting the coefficients via FS(FS-Guided)}
% 	\label{tab:: ablation_adapter}
% 	\begin{tabular}{c|ccc}
% 		\hline
% 		Methods                             & Base Accuracy & New Accuracy &  HM \\
% 		\hline
% 		No Adapter   &         83.69\%         &       74.91\%          &           78.74\%               \\
% 		Fixed   &         \textbf{84.20\%}         &       75.26\%          &           79.02\%                 \\
% 		FS Fixed &         83.80\%         &      75.01\%           &       78.96\%    \\
% 		FS-Guided &       83.93\%           &     \textbf{75.60\%}            &      \textbf{79.55\%}      \\
		
% 		\hline
% 	\end{tabular}
% \end{table}

\section{Conclusion}

We investigate the reasons behind the degradation of generalization for prompt tuning of vision-language models.
We find that cross-modal misalignment can be quantified with our proposed feature shift.
% for both single-modal and multi-modal 
% in feature shift across different modalities during prompt tuning is the primary cause.
% To address the issue of feature misalignment that arises during the prompt tuning process of the vision-language model,
The inter-modal discrepancy of feature shift is negatively associated with performance gains on both base and novel classes in downstream tasks.
Therefore,
we propose RESTORE to adapt prompts in various modalities via the feature shift consistency loss.
Additionally, we propose the "surgery" block,
a feature-shift guided adapter,
to tackle the potential hacking risk out of overfitting.
% feature shift observed in the backbone of VLMs.
Such adapters effectively adjust representations that undergo large-scale feature shifts.
% have already undergone a shift.
% To assess the effectiveness of our approach, we conducted extensive experiments, comparing our methods against existing approaches in three important evaluation settings.
Extensive experiments demonstrate that our method surpasses SOTA methods under multiple evaluation settings.
% in performance.

There are mainly two drawbacks associated with the proposed method.
First,
it might not be accurate enough to use the mean squared error (MSE) and the Frobenius norm for measuring the discrepancy between feature shifts from different modalities.
% may not provide accurate enough measurements to
Alternative distance or divergence measures might be more constructive in capturing such discrepancy.
% the distance between features of different modalities.
Second,
the exploration of prompt tuning might be more focused on engineering and proper theoretical analysis of the relationship between the degenerated model and biased prompts is still lacking.
In our future work,
we plan to 1) propose evaluation tools of overfitting for prompt tuning with theoretical support,
and 2) validate the proposed method on models with larger sizes and generative multi-modal models.

\bibliographystyle{splncs04}
\bibliography{main}
\end{document}